\documentclass[letterpaper]{article} 
\usepackage{aaai25}  
\usepackage{times}  
\usepackage{helvet}  
\usepackage{courier}  
\usepackage[hyphens]{url}  
\usepackage{graphicx} 
\usepackage{natbib}  
\usepackage{caption} 
\usepackage{algorithm}
\usepackage{algorithmic}
\usepackage{multirow}
\usepackage{booktabs}
\usepackage{amssymb}
\usepackage{amsmath}
\usepackage{colortbl}
\definecolor{lightblue}{rgb}{0.93,0.95,1.0}

\usepackage{newfloat}
\usepackage{listings}
\DeclareCaptionStyle{ruled}{labelfont=normalfont,labelsep=colon,strut=off} 
\floatstyle{ruled}
\newfloat{listing}{tb}{lst}{}
\floatname{listing}{Listing}
\title{Wills Aligner: Multi-Subject Collaborative Brain Visual Decoding}
\author{
    Guangyin Bao\textsuperscript{\rm 1},
    Qi Zhang\textsuperscript{\rm 1},
    Zixuan Gong\textsuperscript{\rm 1},
    Jialei Zhou\textsuperscript{\rm 1}, \\
    Wei Fan\textsuperscript{\rm 2}, 
    Kun Yi\textsuperscript{\rm 3}, 
    Usman Naseem\textsuperscript{\rm 4}, 
    Liang Hu\textsuperscript{\rm 1}, 
    Duoqian Miao\textsuperscript{\rm 1}\thanks{Corresponding author.}
}
\affiliations{
    \textsuperscript{\rm 1}Tongji University\\
    \textsuperscript{\rm 2}University of Oxford\\
    \textsuperscript{\rm 3}North China Institute of Computing Technology, State Information Center of China\\
    \textsuperscript{\rm 4}Macquarie University\\
    \{baogy, zhangqi\_cs, gongzx, dqmiao\}@tongji.edu.cn
}

\begin{document}

\maketitle

\begin{abstract}
Decoding visual information from human brain activity has seen remarkable advancements in recent research. However, the diversity in cortical parcellation and fMRI patterns across individuals has prompted the development of deep learning models tailored to each subject. The personalization limits the broader applicability of brain visual decoding in real-world scenarios. To address this issue, we introduce Wills Aligner, a novel approach designed to achieve multi-subject collaborative brain visual decoding. Wills Aligner begins by aligning the fMRI data from different subjects at the anatomical level. It then employs delicate mixture-of-brain-expert adapters and a meta-learning strategy to account for individual fMRI pattern differences. Additionally, Wills Aligner leverages the semantic relation of visual stimuli to guide the learning of inter-subject commonality, enabling visual decoding for each subject to draw insights from other subjects' data. We rigorously evaluate our Wills Aligner across various visual decoding tasks, including classification, cross-modal retrieval, and image reconstruction. The experimental results demonstrate that Wills Aligner achieves promising performance.
\end{abstract}

\section{Introduction}
The process of human perception is marvelous. 
Our perception of the world is shaped not only by the objective reality around us but also by our individual subjective experiences.
Understanding the mechanisms of human perception is crucial for unraveling the complexities of the brain, advancing brain-inspired computational models~\cite{pami,brain_decoding_1,brain_decoding_2,brain_inMM1}, and offering numerous applications in clinical medicine~\cite{fmribook} and brain-computer interfaces~\cite{brain_decoding_11, brain_decoding_21}.
Within this expansive field, visual decoding stands out as a critical and challenging study. It enables us to delve into the intricate workings of the brain during visual processing, object recognition, and scene interpretation~\cite{brain_visual_1,brain_visual_2}. 
Among the various brain imaging modalities, functional magnetic resonance imaging (fMRI) is particularly favored by researchers due to its non-invasive nature and its ability to precisely localize functional regions of the cortex. Consequently, fMRI-based visual decoding has become a significant and prominent topic in neuroscience research.

\begin{figure}[!t]
  \centering
  \includegraphics[width=0.98\linewidth]{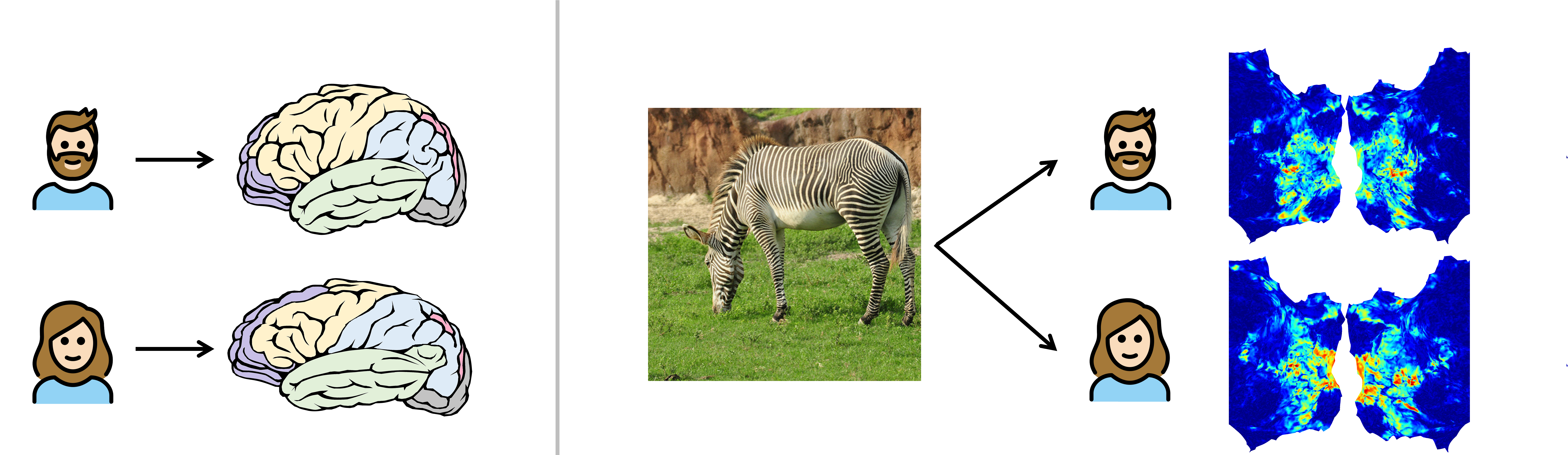}
   \caption{Illustrations of inter-subject brain differences.
   The left figure illustrates structural differences, showing variations in size and location of the same functional areas (highlighted in the same color) across subjects. The right figure depicts differences in fMRI patterns, where distinct brain activities are observed when the same stimulus is presented.}
   \label{fig: introduction}
\end{figure}

Extensive studies~\cite{god,mindreader,braindiffuser,mindeye,brainclip} have investigated brain visual decoding. They construct classification, retrieval, and image reconstruction tasks to mine visual information from fMRI. However, these methods encounter significant challenges in practical application, primarily because they are customized for individual subjects. 
To elaborate, a subject-specific deep model is developed by utilizing the subject's fMRI data, necessitating an expanding number of models due to the substantial differences in brains.
These differences arise from variations in brain structure and fMRI patterns, attributed to diverse genetic backgrounds and cognitive development processes. As shown in Figure~\ref{fig: introduction}, structural differences pertain to variations in cortical parcellation among subjects, with disparities in the size and location of functional areas like visual, language, and memory regions. It can only be identified through anatomical information such as gray and white matter~\cite{fsaverage, cortical}. The fMRI pattern differences refer to the fact that different subjects exhibit distinct brain activity even when exposed to the same visual stimulus. These factors complicate the development of multi-subject visual decoding models. Hence, it is crucial to explore feasbile approaches that can achieve effective collaborative visual decoding across subjects while accommodating individual differences.

Currently, research on multi-subject collaboration in visual decoding is quite limited, and existing multi-subject models often underperform compared to single-subject models. Addressing structural differences between subjects is a prerequisite for effective collaboration. Recent methods~\cite{clip-mused, mindbridge} tackle this challenge with basic data preprocessing techniques like PCA or max pooling, which, however, can lead to brain information loss and misalignment of functional regions. Other approaches~\cite{througheyes} achieve fMRI structural alignment by training linear models based on the same images viewed by different subjects. These methods also struggle with scalability due to the need for identical visual stimuli across all subjects. Fortunately, advances in neuroscience offer inspiration by leveraging prior anatomical knowledge and aligning fMRI data to a standardized brain template.
Besides achieving fMRI alignment, another challenge is to perform effective multi-subject collaboration. Existing approaches either employ subject-specific tokens to identify each subject~\cite{clip-mused} or adapt the initial ridge regression to individual subjects~\cite{scotti2024mindeye2}, allowing model training to accommodate individual differences.
Although making desirable progress, these methods only achieve superficial subject information integration from the data input aspect rather than deep semantics of representations.
In addition, they have primarily focused on handling fMRI pattern differences while overlooking inter-subject commonality, resulting in visual decoding that benefits only from data within a single subject rather than multiple subjects. Naturally, it is necessary to learn both the inter-subject commonality and perceive various fMRI patterns for achieving multi-subject collaboration.

In light of the above discussion, we propose Wills Aligner. We employ an fMRI alignment technique derived from neuroscience, termed anatomical alignment, to address brain structural differences.
To facilitate effective multi-subject collaboration, we first employ the semantic relation of visual stimuli to guide the model in learning inter-subject commonality, thereby enabling the transfer of universal fMRI visual decoding knowledge among subjects. Following this, we introduce Mixture-of-Brain-Expert (MoBE) adapters, which are subject-guided sparse MoE networks designed to capture distinct fMRI patterns. Additionally, we implement a meta-learning strategy that progressively integrates these learned fMRI patterns into the semantics of deep representations, enhancing multi-subject decoding performance.

We conducted a comprehensive evaluation on the Natural Scene Dataset (NSD)~\cite{nsd}, including multi-label classification, bidirectional retrieval, few-shot learning, and fMRI-to-image reconstruction. Our experimental results demonstrate that the Wills Aligner consistently delivers promising performance, surpassing both single-subject and multi-subject baselines under equivalent conditions. 

Our contributions are summarized as follows:
\begin{itemize}
\item We identify key limitations in existing visual decoding methods and propose a multi-subject collaborative approach that addresses these challenges.
\item We utilize the anatomical alignment to address structural differences in fMRI data across subjects, experimentally demonstrating its superiority over alternative fMRI preprocessing methods.
\item We leverage the semantic relation of visual stimuli to guide inter-subject commonality learning and introduce MoBE adapters with a meta-learning strategy to capture different fMRI patterns.
\end{itemize}

\begin{figure*}[t]
  \centering
  \includegraphics[width=0.98\linewidth]{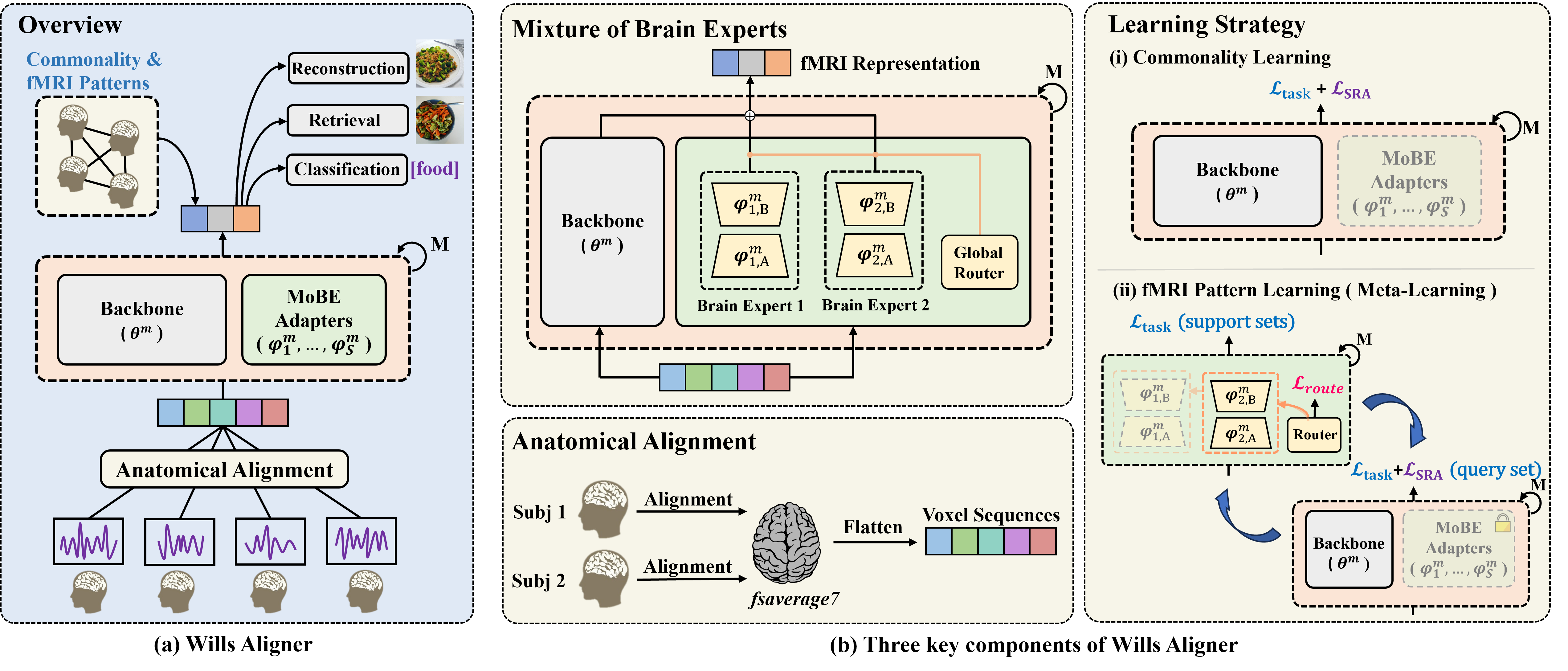}
   \caption{Overview of the proposed Wills Aligner. Figure (a) illustrates the pipeline of our method. Figure (b) shows its three key components: Anatomical Alignment, Mixture of Brain Experts, and Learning Strategy.}
   \label{fig: pipeline}
\end{figure*}

\section{Related Works}
\subsection{fMRI Visual Decoding}
\label{related work: fMRI visual decodeing}

The study of brain visual decoding based on fMRI has been a long-standing endeavor~\cite{minddiffuser, mindeye, scotti2024mindeye2, mindtuner, neuroclips, ren2025improving, huo2025neuropictor, li2025enhancing}.
Due to the low signal-to-noise ratio of fMRI, early studies~\cite{takagi2023high} emphasized the use of purely linear models, such as linear regression, to embed fMRI and images into a common intermediate space.
Further studies~\cite{pami, god, shen2019deep, mindreader} used VGG or ResNet to extract semantically rich image representations.
Mindeye~\cite{mindeye} demonstrated that a large MLP can serve as an excellent brain representation learner.
Other work~\cite{clip-mused,mind-vis,mindvideo,lite-mind} attempted to use Transformer-style~\cite{transformer} in brain visual decoding.
As for optimization objectives, early studies~\cite{god,fmri_cls} used the cross-entropy of categorization tasks as supervised objectives. 
Recognizing visual stimuli categories from fMRI can decode coarse-grained visual information.
Mind-Reader~\cite{mindreader} first used an fMRI-to-image retrieval task to decode fine-grained visual information.
Inspired by contrastive learning~\cite{clip}, Brain-Diffuser~\cite{braindiffuser} aligned fMRI representations to pre-trained ViT's latent space.
MindEye~\cite{mindeye} proposed BiMixCo, a data enhancement strategy, to achieve high-performance fMRI-to-image reconstruction.
Some other work explored self-supervised objectives, such as MinD-Vis~\cite{mind-vis} and MindVideo~\cite{mindvideo} used masked brain modeling.
However, Due to the large variation among subjects, the above methods only trained proprietary single-subject models for each subject.

\subsection{Multi-Subject Visual Decoding}
Some advanced studies have explored multi-subject brain visual decoding. 
They focus on how to align fMRI data from various subjects and how to learn inter-subject differences.
CLIP-MUSED~\cite{clip-mused} first studies multi-subject visual classification, and its performance reaches the level of a single-subject model.
MindBridge~\cite{mindbridge} aligns fMRI using adaptive fMRI aggregation.
Psychometry~\cite{psychometry} learns inter-subject differences using dense MoE with subject-specific projector layers.
MindEye2~\cite{scotti2024mindeye2} aligns fMRI using simple subject-specific ridge regressions.
Although these methods achieve better decoding performance than the single-subject methods, they ignore learning inter-subject commonalities.

\section{Method}

\subsection{Formulation and Overview}
\label{method: overview}

Our target is to decode the visual information contained in the fMRI from a total of $S$ subjects. 
Each fMRI dataset can be denoted as $D_s = \{ (X_{s,i}, Y_{s,i}, L_{s,i}, I_{s, i}) \}_{i=1}^{N_s}, s \in \{1,...,S\}$, where $X_{s, i} \in \mathbb{R}^{L_s \times H_s \times W_s}$ represents fMRI, $Y_{s, i}$ is the seen image, $L_{s, i}$ is the image label, and $I_{s, i}$ is subject identity in the form of one-hot encoding.
We explore visual decoding at three distinct levels: recognizing visual categories from fMRI data, performing bidirectional retrieval between fMRI and images, and reconstructing the seen image based on the corresponding fMRI. 

As illustrated in Figure~\ref{fig: pipeline}, our Wills Aligner comprises three key components: 1) \textbf{Anatomical Alignment}, an fMRI preprocessing to handle fMRI structural differences. 2) \textbf{Mixture of Brain Experts}, a subject-guided sparse mixture-of-expert (MoE) network for learning various fMRI patterns. 3) \textbf{Learning Strategy}, a two-phase training strategy to achieve multi-subject effective collaboration.

\subsection{Anatomical Alignment}
The brain structural differences among subjects lead to inconsistency in voxel counts and correspondence. Addressing this issue is a prerequisite for training with fMRI data from multiple subjects.
To uniformly embed fMRI, we introduce the prior anatomical knowledge to assist fMRI alignment, i.e. anatomical alignment.
The anatomical alignment process begins with the construction of the \textit{fsaverage} standard brain template, a triangular surface mesh generated through spherical registration of 40 individual brain structures, based on the gray and white matter distributions, using an energy minimization algorithm.
Then, a well-established mapping function maps any fMRI data to this template, preserving the cortical topological structure as much as possible, formally defined as $f: \mathbb{R}^{L_s \times H_s \times W_s} \rightarrow \mathbb{R}^{H_0 \times W_0}$, where $H_0 \times W_0$ represents the size of the standard brain template \textit{fsaverage7}~\cite{fsaverage}.
Due to the complete retention of brain activity and structure, this alignment strategy can achieve visual information losslessness.
Subsequently, the one-dimensional sequence $\tilde{X}_{s, i} \in \mathbb{R}^{d}$ is obtained through the implementation of anatomical alignment, the selecting a region-of-interest (ROI), and the application of spatial flattening.
After preprocessing, the aligned fMRIs are structure-standardized, but different fMRI patterns remain.

\subsection{Mixture of Brain Experts}
The fMRI pattern differences are the key factor hindering multi-subject collaboration.
To allow one deep model to learn various fMRI patterns simultaneously, we introduce the Mixture of Brain Expert (MoBE) adapters. This branch network can recognize fMRI patterns and then learn using the corresponding brain experts. 
We parameterize the model $\mathcal{F}(\cdot)$ into three parts: $\theta$ is the subject-shared backbone, $\varphi_s$ is MoBE adapters of Subject $s$, and $r$ is the router. 

\subsubsection{MoBE adapters}
Similar to LoRA~\cite{lora}, MoBE adapters $\varphi_s$ use a parallel structure connected to subject-shared parameters $\theta$.
Assuming that the model contains a total of $M$ linear layers, thus there are $\theta = \{\theta^m\}_{m=1}^M $ and $\varphi_s = \{ \varphi_s^m \}_{m=1}^M $. 
We use $x^m$ to denote the input of the linear layer $m$ and $x^1 = \tilde{X}_{s, i}$. 
Then, the output $o^m$ of linear layer $m$ can be formalized as:
\begin{equation}
    o^m =(\theta^m + \sum\limits_{s=1}^S \omega_s \varphi_s^m)~x^m~,
\end{equation}
where $\omega_s$ represents the routing coefficient assigned to each MoBE adapter.
To reduce the number of parameters in the MoBE adapters, we apply a low-rank decomposition to the parameters $\varphi_s$, similar to LoRA. In addition, we adopt the same initialization strategy as LoRA.

\subsubsection{Subject-Guided Sparse Global Router}
Unlike existing MoE architectures that use separate routers for each MoE layer, we configure a subject-guided sparse global router.
We find that the preprocessed fMRI data $\tilde{X}_{s, i}$ retains identifiable subject-specific fMRI patterns, allowing a simple model to classify. 
Therefore, we use subject identity $I_{s, i}$ as supervision, training an MLP to function as the router $\mathcal{R}(\cdot)$: 
\begin{equation}
    \mathcal{L}_{\mathrm{router}} = \sum_{s=1}^S \sum_{i=1}^{N_s} \mathrm{CrossEntropy}(\mathcal{R}(\tilde{X}_{s, i};r), I_{s, i})~.
\end{equation}
In our multi-subject visual decoding tasks, all MoBE adapters must act consistently and make the same routing choices when processing a specific subject's fMRI data. To avoid severe parameter redundancy, we configured a single router to handle routing for all MoBE adapters, i.e. a global router. To ensure the proper functioning of MoBE adapters, it is essential to implement a sparse routing strategy. we use the probability distributions output by the trained router as the routing coefficient:
\begin{equation}
    [ \omega_1, \omega_2, ..., \omega_S ] = \mathcal{R}(\tilde{X}_{s, i};r)~.
\end{equation}
Since the router/classifier can give extremely high classification confidence, this dense routing algorithm is equivalent to a sparse one. We will verify this in our experiments.

\subsection{Learning Strategy}
We split model training into two phases. 
In the first phase, we aim to guide the model backbone $\theta$ in learning the universal fMRI visual decoding knowledge. It is similar to a model pre-training and we achieve this commonality learning by a meticulously designed supervised objective.
In the second phase, we aim to learn various fMRI patterns using MoBE adapters $\varphi_s$. We further employ a meta-learning strategy to integrate the learned fMRI patterns into semantics of deep representations.
The supervised objective of downstream tasks is interspersed throughout both phases.

\subsubsection{Downstream Task Learning}

We use $\mathcal{L}_{\mathrm{task}}$ to represent downstream task supervision. It takes different forms depending on the visual decoding task.
We calculate the fMRI representation $f_{s, i} = \mathcal{F}(\tilde{X}_{s, i};\theta, \varphi_1, ..., \varphi_S, r)$.

For the classification task, we employ a classifier $\mathcal{G}(\cdot)$ and construct loss as follow:
\begin{equation}
    \mathcal{L}_{\mathrm{task}} = \mathcal{L}_{\mathrm{cls}} = 
    \sum_{s=1}^S \sum_{i=1}^{N_s} 
 \mathrm{CrossEntropy}(\mathcal{G}(f_{s,i}), L_{s, i}).
\end{equation}

For the retrieval task, we first use the pre-trained ViT to extract the image representation $y_{s, i} = \mathrm{ViT}(Y_{s, i})$. Then, we use a retrieval projector $\mathcal{H}(\cdot)$ to construct the bidirectional contrastive loss as follow:
\begin{equation}
    \begin{aligned}
    & \mathcal{L}_{\mathrm{fmri}} = 
    - \sum_{j=1}^{|B|}
    \log \frac{exp(\mathcal{H}(f_j)^\intercal \cdot y_j)}
    {\sum_{k=1}^{|B|} exp(\mathcal{H}(f_j)^\intercal \cdot y_k) }~,
    \\
    & \mathcal{L}_{\mathrm{image}} = 
    - \sum_{j=1}^{|B|}
    \log \frac{exp(y_j^\intercal \cdot \mathcal{H}(f_j))}
    {\sum_{k=1}^{|B|} exp(y_j^\intercal \cdot \mathcal{H}(f_k)) }~,
    \\
    & \mathcal{L}_{\mathrm{task}} = \mathcal{L}_{\mathrm{retri}} = 
    \mathcal{L}_{\mathrm{fmri}} + \mathcal{L}_{\mathrm{image}},
    \end{aligned}
\end{equation}
where $B$ is a mini-batch sampled from entire fMRI.

For the image reconstruction task, we adopt the same basic pipeline as MindEye~\cite{mindeye}. We map fMRI representations to the ViT image space through a diffusion prior $\mathcal{D}(\cdot)$. The reconstruction loss contains contrastive loss:
\begin{equation}
     \mathcal{L}_{\mathrm{task}} = \mathcal{L}_{\mathrm{recon}} = \mathcal{L}_{\mathrm{prior}}(\mathcal{D}(f_{s,i}), y_{s,i}) + \mathcal{L}_{\mathrm{retri}}~.
\end{equation}

\subsubsection{Commonality Learning} 

Inter-subject commonality is universal knowledge used for fMRI visual decoding, which is supposed to be independent of subject identity. Essentially, it implies one-to-one correspondences between fMRI and images. Therefore, a natural insight is to increase the relevance of fMRI to the image and weaken its relevance to the subject's identity, which can be achieved through contrastive learning. To construct contrastive pairs, we utilize the similarity relation of representations.
We represent fMRI representations, image representations, and subject identities in a mini-batch by $\mathbb{F}$, $\mathbb{Y}$, and $\mathbb{I}$, respectively.
Then, we calculate their respective similarity relations by
$\mathcal{M}_F = \mathbb{F}^\intercal\mathbb{F}$,
$\mathcal{M}_Y = \mathbb{Y}^\intercal\mathbb{Y}$, 
and $\mathcal{M}_I = \mathbb{I}^\intercal\mathbb{I}$.
Subsequently, We align the fMRI relation more closely with the image relation while distancing it from the subject identity relation, and term this process Semantic Relation Alignment (SRA):
\begin{equation}
    \mathcal{L}_{\mathrm{SRA}} = - \log 
    \frac { \mathrm{sim}(\mathcal{M}_F, \mathcal{M}_Y) }
    { \mathrm{sim}(\mathcal{M}_F, \mathcal{M}_Y) + \mathrm{sim}(\mathcal{M}_F, \mathcal{M}_I) }~,
\end{equation}
where $\mathrm{sim}(\cdot)$ represents the cosine similarity. Based on this, our first training phase uses task loss and SRA loss to supervise the model backbone $\theta$ in learning commonality:
\begin{equation}
    \theta^* = \arg\min_{\theta} \sum_{s=1}^S \sum_{i=1}^{N_s} (\mathcal{L}_{\mathrm{task}} + \alpha \mathcal{L}_{\mathrm{SRA}})~.
\end{equation}
Here $\alpha$ is the factor to balance two losses.

\begin{table*}[t]
\centering
    \centering
  \scalebox{0.95}{
  \begin{tabular}{@{}ccccccc@{}}
    \toprule
    Method & \# Model & \# Parameters & mAP$\uparrow$ & AUC$\uparrow$ & Hamming$\downarrow$ \\
    \midrule
     Single-Subject Vanilla Method            &  4    & 66M         &  0.258  &  0.854  &  0.033  \\
     Multi-Subject Vanilla Method             &  1    & 66M        &  0.150  &  0.767  &  0.039  \\
     EMB~\cite{emb}            &  1   &  66M  &  0.220  &  0.825  &  0.035  \\
     CLIP-MUSED~\cite{clip-mused}&  1   &  66M &  0.258  &  0.877  &  0.030  \\
     \rowcolor{lightblue}
     Wills Aligners (\textbf{ours})      &  1  &  \textbf{19M}           &  \textbf{0.424}  &  \textbf{0.937}  &  \textbf{0.024}  \\
    \bottomrule
  \end{tabular}
  }
  \caption{Experimental results of the multi-label classification task. All baseline results are quoted from CLIP-MUSED~\cite{clip-mused}. We report the average value of four subjects, and our results are averaged over three runs.}
  \label{table: classification result}
\end{table*}

\begin{table*}[t]
    \centering
    \scalebox{0.95}{
    \begin{tabular}{@{}ccccc@{}}
    \toprule
    \multirow{2}{*}{Method} & \multirow{2}{*}{Objectives} & \multirow{2}{*}{\# Parameters} & \multicolumn{2}{c}{Retrieval Accuracy} \\
    \cline{4-5}
    & & & Image $\uparrow$ & fMRI $\uparrow$  \\
    \midrule
     Mind Reader~\cite{mindreader} &  InfoNCE Loss              &  2M    &  11.0\%  &  49.0\%  \\
     Brain Diffuser~\cite{braindiffuser} &  Contrastive Loss    &  3B    &  29.9\%  &  21.4\%  \\ 
     MindEye~\cite{mindeye}   &  Contrastive Loss                         &  996M  &  83.7\%  &  79.1\%  \\
     MindEye~\cite{mindeye}  &  Contrastive Loss + MSE loss              &  996M  &  88.8\%  &  84.9\%  \\
     MindEye~\cite{mindeye}  &  SoftCLIP Loss + BiMixCo                &  996M  &  89.6\%  &  82.2\%  \\
     \rowcolor{lightblue}
     Wills Aligners (\textbf{ours})   &  Contrastive Loss                         &  \textbf{523M}  &  \textbf{95.4\%}  &  \textbf{83.9\%}  \\
     \bottomrule
    \end{tabular}
  }
  \caption{Experimental results of the cross-modal retrieval on NSD. Our results are averaged over three runs. We report the results on Subj01, as previous works only provided results on Subj01.} 
  \label{table: retrieval result}
\end{table*}

\subsubsection{fMRI Pattern Learning}
We use a meta-learning strategy to learn various fMRI patterns and integrate them into semantics of deep representations. 
Formally, we define the fMRI data of each subject $\{D_s\}_{s=1}^S$ as support sets, and entire fMRI $D=\cup_{s=1}^S D_s$ as the query set.
Then, the bi-level optimization of meta-learning can be formalized as:
\begin{equation}
\begin{aligned}
     \theta^*\! = & \arg\min_{\theta}\! \sum_{\tilde{X} \in D} \!(\mathcal{L}_{\mathrm{task}}+\alpha\mathcal{L}_{\mathrm{SRA}}) (\tilde{X}; \theta, \varphi_1^*, ..., \varphi_S^*),\\
    & \mathrm{s.t.} \quad
    \varphi_s^* = \arg\min_{\varphi_s} \sum_{\tilde{X} \in D_s} \mathcal{L}_{\mathrm{task}} (\tilde{X}; \varphi_s^*)~.
\end{aligned}
\end{equation}
The alternate method is more conducive to the model learning fMRI patterns. During each training session on support sets, MoBE adapters capture fMRI patterns. Subsequent fine-tuning on the query set integrates these fMRI patterns into the backbone network, which achieves improved deep representations and gives back fMRI pattern learning. Repeated execution of this alternating training will gradually improve multi-subject visual decoding performance.

\section{Experiment}
To validate the effectiveness and generalization of our Wills Aligner, we conduct extensive experiments, including classification, retrieval, reconstruction, and few-shot learning on the NSD dataset. Additionally, we perform further evaluations and ablation studies; please refer to our appendix\footnote{https://arxiv.org/abs/2404.13282}.

\subsection{Dataset}
The Natural Scenes Dataset (NSD) is a massive 7T neuroscience dataset encompassing fMRI data.
Throughout the NSD experiment, participants were presented with images sourced from MSCOCO~\cite{mscoco}, while their neural responses were recorded.
Our study aligns with previous research by concentrating on Subj01, Subj02, Subj05, and Subj07, as these 4 subjects completed all 40 session scans.

\subsection{Classification Experiment}
The classification task stands as a coarse-grained brain visual decoding, which demands extracting and recognizing the categories of visual stimuli contained within fMRI data.
We compare Wills Aligner with existing multi-subject classification methods on the NSD dataset. The implementation details follow CLIP-MUSED~\cite{clip-mused}.
This task is a multi-label classification task, we employ three commonly used evaluation metrics in this field: mean Average Precision (mAP), the area under the receiver operating characteristic curve (AUC), and Hamming distance.

\subsubsection{Results and Analysis}

Table~\ref{table: classification result} shows the results of the classification experiment.
The results indicate that our method achieves state-of-the-art performance in coarse-grained brain visual decoding, with a 64.3\% improvement in mAP, a 6.8\% improvement in AUC, and a 20.0\% reduction in Hamming distance, while using fewer model parameters.
The improvement can be attributed to two perspectives. On the one hand, our anatomical alignment is better than the baselines' fMRI alignment (such as PCA). Since their fMRI alignment focuses solely on the value itself without considering the brain structure, it tends to lose valuable information. In contrast, our fMRI alignment is based on the brain's anatomical knowledge, which retains useful brain information. On the other hand, our method achieves multi-subject collaboration so that visual decoding for the current subject benefits from other subjects' data.
The similar performance between the single-subject vanilla method and CLIP-MUSED also suggests that existing methods merely avoid the damage to the performance caused by various fMRI patterns rather than exploiting the benefits of inter-subject commonality. Nevertheless, our method can achieve this.

\begin{table*}[t]
  \centering
    \centering
      \scalebox{0.83}{
  \begin{tabular}{@{}c|c|cccccccccccc@{}}
    \toprule
    \multirow{2}{*}{Few-Shot Ratio} & \multirow{2}{*}{Method} & & \multicolumn{2}{c}{Subject 1} & & \multicolumn{2}{c}{Subject 2} & & \multicolumn{2}{c}{Subject 5} & & \multicolumn{2}{c}{Subject 7} \\
    \cline{4-5}
    \cline{7-8}
    \cline{10-11}
    \cline{13-14}
    & & & mAP $\uparrow$ & AUC $\uparrow$ & & mAP $\uparrow$ & AUC $\uparrow$ & & mAP $\uparrow$ & AUC $\uparrow$ & & mAP $\uparrow$ & AUC $\uparrow$ \\
    \midrule
    0.05        & Vanilla & & 0.128 & 0.782 & & 0.138 & 0.740 & & 0.154 & 0.804 & & 0.122 & 0.779 \\
    (1 session) & Wills Aligner (\textbf{ours})  & & \textbf{0.270} & \textbf{0.901} & & \textbf{0.229} & \textbf{0.880} & & \textbf{0.275} & \textbf{0.907} & & \textbf{0.206} & \textbf{0.862} \\
    \midrule
    0.1         & Vanilla & & 0.143 & 0.808 & & 0.168 & 0.834 & & 0.171 & 0.827 & & 0.138 & 0.802 \\
    (2 sessions)& Wills Aligner (\textbf{ours})  & & \textbf{0.322} & \textbf{0.915} & & \textbf{0.287} & \textbf{0.898} & & \textbf{0.334} & \textbf{0.921} & & \textbf{0.276} & \textbf{0.888} \\
    \midrule
    0.2         & Vanilla & & 0.210 & 0.874 & & 0.185 & 0.856 & & 0.179 & 0.851 & & 0.164 & 0.830 \\
    (4 sessions)& Wills Aligner (\textbf{ours}) & & \textbf{0.385} & \textbf{0.930} & & \textbf{0.354} & \textbf{0.922} & & \textbf{0.409} & \textbf{0.935} & & \textbf{0.317} & \textbf{0.904} \\
    \bottomrule
  \end{tabular}
  }
  \caption{Experiment results of few-shot classification on NSD.} 
  \label{table: few-shot classification}
\end{table*}

\begin{table*}[t]
  \centering
  \scalebox{0.78}{
  \begin{tabular}{@{}c|c|cccccccccccc@{}}
    \toprule
    \multirow{2}{*}{Few-Shot Ratio} & \multirow{2}{*}{Method} & & \multicolumn{2}{c}{Subject 1} & & \multicolumn{2}{c}{Subject 2} & & \multicolumn{2}{c}{Subject 5} & & \multicolumn{2}{c}{Subject 7} \\
    \cline{4-5}
    \cline{7-8}
    \cline{10-11}
    \cline{13-14}
    & & & Image $\uparrow$ & fMRI $\uparrow$ & & Image $\uparrow$ & fMRI $\uparrow$ & & Image $\uparrow$ & fMRI $\uparrow$ & & Image $\uparrow$ & fMRI $\uparrow$ \\
    \midrule
    0.05        & Vanilla & & 10.6\% & 1.9\% & & 11.8\% & 1.2\% & & 10.7\% & 1.8\% & & 9.1\% & 1.2\% \\
    (1 session) & Wills Aligner (\textbf{ours})  & & \textbf{65.1\%} & \textbf{47.9\%} & & \textbf{69.9\%} & \textbf{49.3\%} & & \textbf{47.0\%} & \textbf{29.3\%} & & \textbf{46.3\%} & \textbf{30.0\%} \\
    \midrule
    0.1         & Vanilla & & 26.5\% & 4.6\% & & 29.5\% & 3.5\% & & 20.0\% & 3.2\% & & 20.9\% & 2.6\% \\
    (2 sessions)& Wills Aligner (\textbf{ours})  & & \textbf{75.9\%} & \textbf{53.0\%} & & \textbf{77.4\%} & \textbf{56.3\%} & & \textbf{53.8\%} & \textbf{35.6\%} & & \textbf{54.8\%} & \textbf{39.4\%} \\
    \midrule
    0.2         & Vanilla & & 46.3\% & 13.8\% & & 53.8\% & 12.8\% & & 43.9\% & 8.5\% & & 41.1\% & 8.6\% \\
    (4 sessions)& Wills Aligner (\textbf{ours})  & & \textbf{82.7\%} & \textbf{61.1\%} & & \textbf{84.2\%} & \textbf{64.2\%} & & \textbf{64.3\%} & \textbf{46.6\%} & & \textbf{63.0\%} & \textbf{46.8\%} \\
    \bottomrule
  \end{tabular}
  }
  \caption{Experiment results of few-shot retrieval on NSD.} 
  \label{table: few-shot retrieval}
\end{table*}

\begin{table*}[t]
  \centering
  \scalebox{0.83}{
  \begin{tabular}{@{}cccccccccccc@{}}
    \toprule
    \multirow{2}{*}{Methods}&\multirow{2}{*}{\# Models}&\multicolumn{4}{c}{Low-Level}&&\multicolumn{4}{c}{High-Level}\\
    \cline{3-6}
    \cline{8-11}
&&PixCorr$\uparrow$&SSIM$\uparrow$&Alex(2)$\uparrow$&Alex(5)$\uparrow$&&Incep$\uparrow$&CLIP$\uparrow$&Eff$\downarrow$&SwAV$\downarrow$\\
    \midrule
Mind-Diffuser~\cite{minddiffuser}&4&0.254&\textbf{0.356}&94.2\%&96.2\%&&87.2\%&91.5\%&0.775&0.423\\
MindEye~\cite{mindeye}&4&\textbf{0.309}&0.323&94.7\%&\underline{97.8\%}&&93.8\%&94.1\%&\underline{0.645}&\textbf{0.367} \\     DREAM~\cite{xia2024dream}&4&\underline{0.288}&\underline{0.338}&\underline{95.0\%}&97.5\%&&\textbf{94.8\%}&\textbf{95.2\%}&\textbf{0.638}&0.413\\
MindBridge~\cite{mindbridge}&4&0.148&0.259&86.9\%&95.3\%&&92.2\%&94.3\%&0.713&0.413\\
MindBridge~\cite{mindbridge}&\textbf{1}&0.151&0.263&87.7\%&95.5\%&&92.4\%&94.7\%&0.712&0.418\\
     \rowcolor{lightblue} Wills Aligner (\textbf{ours})&\textbf{1}&0.271&0.328&\textbf{95.8\%}&\textbf{98.0\%}&&\underline{94.3\%}&\underline{94.8\%}&0.649&\underline{0.373}\\
    \bottomrule
  \end{tabular}
  }
  \caption{Experiment results on fMRI-to-image reconstruction on NSD. Results are averaged over 4 subjects.}
  \label{table: NSD Reconstruction}
\end{table*}

\subsection{Retrieval Experiment}
The cross-modal retrieval between fMRI and images stands as a fine-grained brain visual decoding task, requiring fine-grained recognition of the visual semantics contained in the fMRI. Following MindEye~\cite{mindeye}, we conduct bidirectional retrieval on NSD.
We evaluate the performance using Top-1 retrieval accuracy. For image retrieval, we compute the cosine similarity between an fMRI representation and its respective ground truth image representation and 299 other randomly selected image representations in the test set.
For each test sample, success is determined if the cosine similarity is greatest between the fMRI representation and its respective ground truth image representations (random chance is 1/300). We repeat the evaluation for each test sample 30 times to account for the variability in the random sampling of batches.
The same procedure is used for fMRI retrieval, except fMRI and images are flipped.

\subsubsection{Results and Analysis}
Retrieval results are shown in Table~\ref{table: retrieval result}.
Our Wills Aligner outperforms MindEye, achieving a 15.4\% performance improvement in image retrieval when the supervised objectives are equal. Even when MindEye uses BiMixCo data augmentation and SoftCLIP loss, it still outperforms MindEye, achieving a 7.8\% improvement, while using fewer model parameters.
Since all baselines are single-subject methods, there is no information loss of fMRI, and differential fMRI patterns are not a concern. Therefore, the results that more accurate visual information is decoded experimentally prove that preprocessed fMRI by anatomical alignment is almost equivalent to the original fMRI without losing visual information.
In Addition, the performance improvement provided by Wills Aligner stems from multi-subject collaboration. Other subjects' data indeed helps fine-grained visual decoding for Subj01.

\begin{table*}[t]
  \centering
  \scalebox{0.80}{
  \begin{tabular}{@{}ccccccccccccccccc@{}}
    \toprule
    \multirow{2}{*}{MoBEs} & \multirow{2}{*}{SRA} & & \multicolumn{2}{c}{Classifiaction} & & \multicolumn{2}{c}{Retrieval} & & \multicolumn{8}{c}{Reconstruction} \\
    \cline{4-5}
    \cline{7-8}
    \cline{10-17}
     & & & mAP $\uparrow$ & AUC $\uparrow$ & & Image $\uparrow$ & fMRI $\uparrow$ & & PixCorr$\uparrow$&SSIM$\uparrow$&Alex(2)$\uparrow$&Alex(5)$\uparrow$&Incep$\uparrow$&CLIP$\uparrow$&Eff$\downarrow$&SwAV$\downarrow$\\
    \midrule
    $\times$ & $\times$         & & 0.314 & 0.904 & & 81.6\% & 77.9\% & & 0.158 & 0.262 & 84.1\% & 93.6\% & 90.8\% & 92.6\% & 0.723 & 0.420\\
    $\checkmark$ & $\times$     & & 0.417 & 0.935 & & 92.3\% & 81.4\% & & \textbf{0.274} & 0.327 & 94.5\% & 96.6\% & 91.3\% & 94.3\% & 0.689 & 0.382\\
    $\times$ & $\checkmark$     & & 0.361 & 0.924 & & 85.8\% & 79.1\% & & 0.143 & 0.255 & 86.7\% & 94.2\% & 92.3\% & 94.4\% & 0.715 & 0.423\\
    \rowcolor{lightblue}
    $\checkmark$ & $\checkmark$ & & \textbf{0.424} & \textbf{0.937} & & \textbf{95.4\%} & \textbf{83.9\%}&&0.271&\textbf{0.328}&\textbf{95.8\%}&\textbf{98.0\%}&\textbf{94.3\%}&\textbf{94.8\%}&\textbf{0.649}&\textbf{0.373}\\
    \bottomrule
    \end{tabular}
    }
  \caption{Results of Ablation experiments. Results are averaged over 4 subjects.} 
  \label{table: ablation}
\end{table*}

\subsection{Few-Shot Visual Decoding Experiment}
We perform the few-shot brain visual decoding experiment to further explore how much a single subject's visual decoding performance can benefit from multi-subject collaboration. 
The few-shot setting is more practical for real-world applications, as collecting large-scale fMRI data (40 sessions) for a single subject is challenging.
Our experiments involve the classification and retrieval on NSD. We employ the few-shot setting for a given subject while the other subjects use the entire fMRI data for training. The few-shot ratios are set to be 0.05, 0.1, and 0.2, corresponding to 1, 2, and 4 sessions of fMRI. 
We compare our Wills Aligner with the single-subject vanilla method, where the model is trained exclusively on fMRI data from the few-shot subjects.

\subsubsection{Results and Analysis}
Tables~\ref{table: few-shot classification} and Table~\ref{table: few-shot retrieval} show the performance of different few-shot subjects in the classification and retrieval tasks.
The experiment results show that our Wills Aligner can significantly improve the brain visual decoding performance of few-shot subjects.
In particular, for Subj01, we achieve a comparable image retrieval performance to MindEye (82.7\% v.s. 83.7\%) while we only use 4-session fMRI (1/10 of MindEye's training data).
This suggests that our multi-subject collaboration strategy can transfer universal visual decoding knowledge from other subjects and improve the decoding performance of the current few-shot subject, highlighting the importance of collaboration.

\subsection{Image Reconstruction Experiment}
We further perform the fMRI-to-image reconstruction task on the NSD dataset to intuitively demonstrate the results of our visual decoding. Following previous studies~\cite{mindeye, mindbridge}, we use eight image quality evaluation metrics. Among these, PixCorr, SSIM, AlexNet(2), and AlexNet(5) assess low-level perceptual aspects, while Inception, CLIP, EffNet-B, and SwAV evaluate high-level semantic aspects.
We compare Wills Aligner with all existing methods. These baselines all used CLIP ViT-L/14 to extract image representations, leading to a fair comparison. 

\subsubsection{Quantitative Results and Analysis}
The quantitative experimental results are shown in Table~\ref{table: NSD Reconstruction}. When compared with the state-of-the-art multi-subject reconstruction method, MindBridge, we achieve performance excellence in all metrics. Such an improvement stems from a better fMRI alignment strategy and multi-subject collaboration. Specifically, the anatomical alignment we use preserves useful fMRI information, while the max pooling employed by MindBridge may result in information loss. The fact that the single-subject MindBridge performs lower than MindEye also confirms the existence of information loss. In addition, our method can capture both inter-subject commonality and different fMRI patterns, which MindBridge lacks.
Even when compared to existing single-subject reconstruction methods, our Wills Aligner also achieves promising performance on some metrics. Since the multi-subject collaboration strategy tends to enhance the model's ability in visual semantic decoding, we fail to achieve a better performance than single-subject baselines on low-level metrics based on pixels, such as PixCorr and SSIM.

\subsubsection{Qualitative Results}
\begin{figure}[t]
  \centering
  \includegraphics[width=0.93\linewidth]{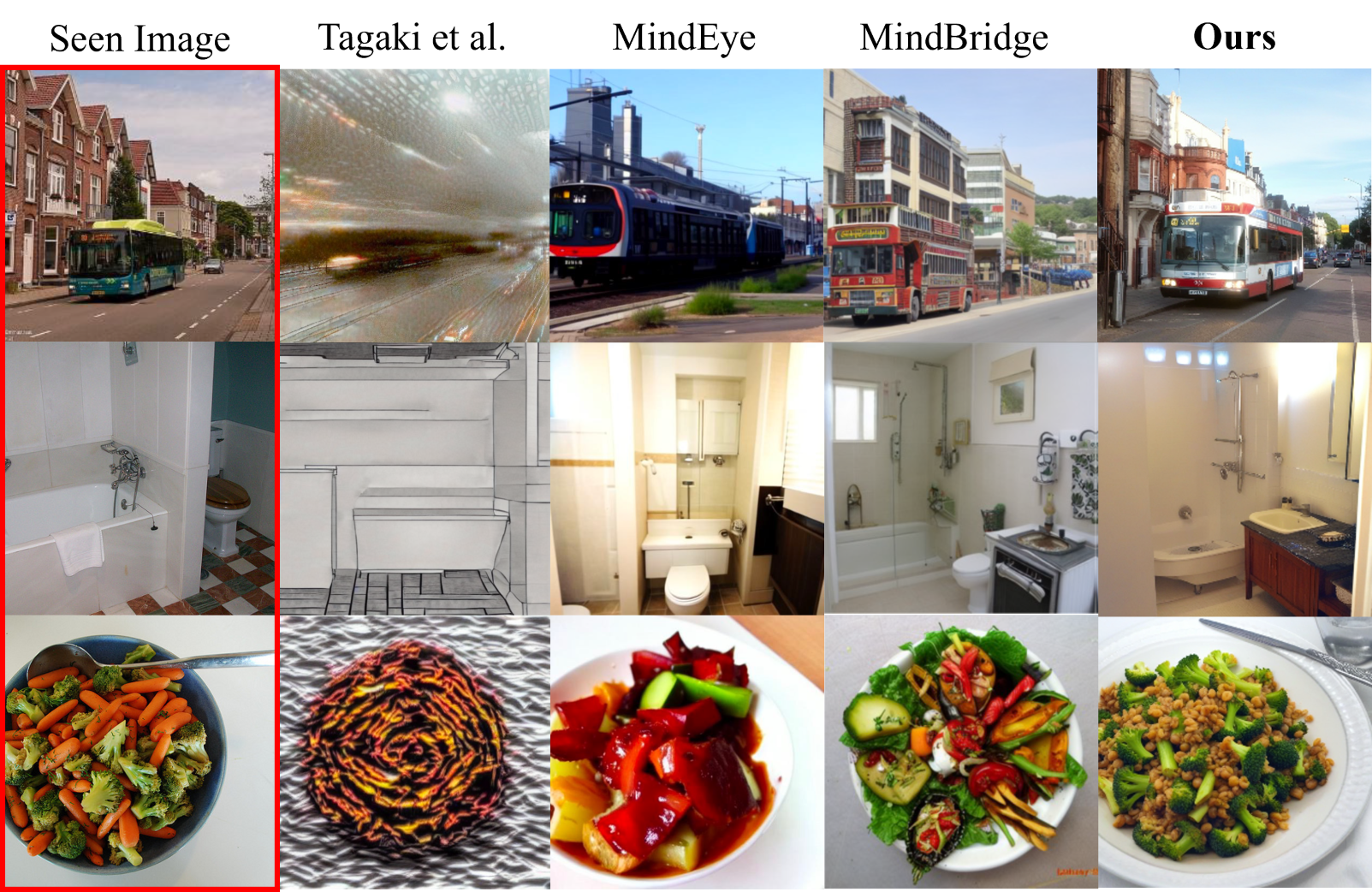}
   \caption{Reconstruction results for Subj01.}
   \label{fig: visualization}
\end{figure}

Figure~\ref{fig: visualization} visualizes the results of our image reconstruction. It can be seen that our reconstruction is more advantageous in terms of semantic accuracy compared to MindBridge. 
For instance, in the top row, our reconstruction of the “bus” exhibits a better spatial position, which is more similar to the seen image. Similarly, in the bottom row, we have accurately reconstructed “broccoli” and “cutlery”.
These observations further explain why our quantitative metrics are better than baselines.

\begin{figure}[t]
  \centering
  \includegraphics[width=0.88\linewidth]{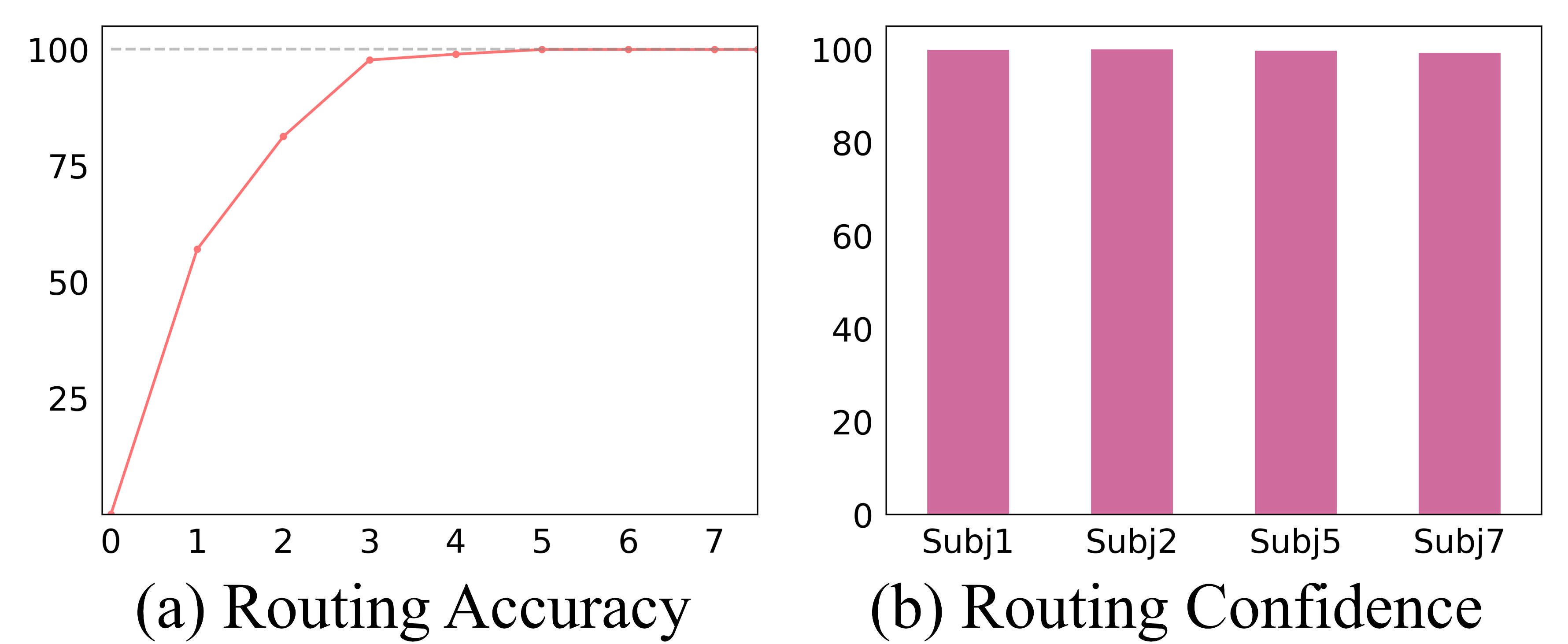}
   \caption{Analysis on router. The left figure shows the growth of the router's test set classification accuracy with training steps. The right figure shows the average probabilities of the router's top-1 prediction on the test set.}
   \label{fig: routing}
\end{figure}

\subsection{Exploratory Experiments}

\subsubsection{Ablation Experiments}
We perform ablation experiments to explore the performance sources of our Wills Aligner.
We ablate two key components of our Wills Aligner.
One is MoBE adapters together with the corresponding meta-learning strategy, and the other is the SRA loss used for learning inter-subject commonality.
The experimental results are shown in Table~\ref{table: ablation}, indicating that both key components of Wills Aligner are useful for performance improvement.
We also find that SRA loss does not improve performance as much as MoBE adapters, suggesting fMRI pattern learning is indispensable in multi-subject collaboration.

\subsubsection{Analysis on Router}
Figure~\ref{fig: routing} illustrates the analysis of the router. It is evident that, despite anatomical alignment, fMRI from different subjects exhibit distinct fMRI patterns, allowing recognition by a simple MLP. After training, the router achieves 100\% accuracy with high confidence, demonstrating that our router achieves sparse routing in fact.

\section{Conclusion}
In this paper, we propose Wills Aligner. It achieves multi-subject collaborative brain visual decoding by anatomical alignment, learning inter-subject commonality, and learning various fMRI patterns.
We have substantiated its effectiveness and generalization through extensive experiments.
Future works could consider fusing fMRI data from different subjects to achieve unbiased visual decoding or use trained decoding models to assist neuroscience research.

\section{Acknowledgement}
This research is supported by the National Key Research and Development Program of China (No.~2022YFB3104700), the National Natural Science Foundation of China (No.~62376198, No.~62076182), Shanghai Baiyulan Pujiang Project (No.~08002360429).

\bibliography{aaai25}

\clearpage

\setcounter{figure}{4}
\setcounter{table}{7}
\setcounter{equation}{9}

\appendix
\section{A.~~Details on fMRI-to-Image Reconstruction}
Our image reconstruction uses a pipeline similar to MindEye~\cite{mindeye}. Here, we first illustrate the details of Diffusion Prior and Low-Level Visual Reconstruction. Then, we illustrate the entire process of image reconstruction.

\subsection{A.1~~Diffusion Prior}
As illustrated in the main text, we use a Diffusion Prior $\mathcal{D}(\cdot)$ to map fMRI representations $f_{s, i}$ to the ViT image space. 

\textbf{Model}. 
The MindEye's Diffusion Prior we adopt is a modified DALLE2 diffusion prior~\cite{ramesh2022dalle2}. Specifically, the inputs for MindEye's Diffusion Prior are 257×768 fMRI representations, 1×768 time embedding, and 257×768 noised image representations. The outputs are 257×768 denoised image representations. Unlike the existing DALLE2 prior, MindEye does not use learnable queries and instead directly predicts denoised image representations from the noised ones. MindEye adds absolute positional embeddings to the noised image representation and uses bidirectional attention rather than causal attention. Moreover, MindEye's Diffusion Proir can work with just 100 timesteps instead of 1000 as used in DALLE2. To achieve multi-subject visual decoding, We modify MindEye's Diffusion Prior with our MoBE adapters.

\textbf{Training}.
Using $d$ to parameterize the Diffusion Prior, its output can be formalized as $\hat{y}_{s, i} = \mathcal{D}(f_{s, i};d)$. The training target is to minimize the mean square error (MSE):
\begin{equation}
    d^* = \arg \min_d ||\hat{y}_{s, i} - y_{s, i}||
\end{equation}

\subsection{A.2~~Low-Level Visual Reconstruction}
The above process of generating predicted image representations $\hat{y}_{s, i}$ based on 1-D voxel sequences $\tilde{X}_{s, i}$ is regarded as the high-level visual reconstruction or semantic reconstruction, and the low-level visual reconstruction is independent of the above process, which aims to map voxel sequences to the latent space of Stable Diffusion’s VAE. 

\textbf{Model}.
To map to Stable Diffusion’s VAE latent space, MindEye uses a separate residual MLP backbone with 4 residual blocks that map 1-D voxel sequences to a 16×16×64 dimensional latent space. Then, a CNN upsamples these results by 4× to create low-level fMRI representations of size (64, 64, 4). The CNN upsampler uses an architecture similar to Stable Diffusion’s VAE decoder. 
To create the targets for the upsampler, MindEye upsamples images to 512×512 through bilinear interpolation and encodes them with the SD VAE encoder. The resulting (64, 64, 4) low-level image representations form the targets for low-level visual reconstruction. 
We modify MindEye's low-level modules with our MoBE adapters to achieve multi-subject reconstruction.

\textbf{Training}.
We use mean absolute error (MAE) between low-level fMRI representations and low-level image representations as supervised objectives of low-level modules. The SoftCLIP loss with another retrieval projector similar to $\mathcal{H}(\cdot)$ is employed in the low-level reconstruction, which acts as an auxiliary loss to improve the performance.

\textbf{Inference}. 
The output of low-level visual reconstruction can be fed into the VAE decoder to produce blurry images that lack semantic content but exhibit state-of-the-art low-level image metrics.

\subsection{A.3~~Image Reconstruction}
The fMRI-to-image reconstruction uses a trained generative model, Versatile Diffusion~\cite{vd}. MindEye adopts the image-to-image transformation of Versatile Diffusion. It takes low-level blurry images and high-level predicted image representation as inputs, thereby reconstructing images.

\section{B.~~Implementation Details}

\subsection{B.1~~Classification on NSD}
\quad \textbf{Model}. 
Consistent with the baseline, we use the Transformer architecture as the model backbone. To allow standard visual transformers to handle one-dimensional voxel sequences, we first perform tokenization to extract voxel tokens, as shown in Appendix C.2. We deploy a 6-layer transformer, use the embedded CLS tokens as fMRI representations, and use a simple classifier for visual classification. The parameter count of the entire model is around 19M.

\textbf{Training}. 
We use the AdamW~\cite{adamw} optimizer. 
The learning rate is searched in 1e-4 using a linear learning rate decay with a final learning rate of 0.01 initial value. We fix weight decay to 1e-4 and set the batch size to 1024.
For commonality learning, We set the weight of SRA loss to 0.1 and train 1000 epochs.
For fMRI pattern learning, we optimize the router for 10 epochs and then optimize the MoBEs for 10 meta steps. Each meta step contains training on the support set for 20 epochs and training on the query set for 5 epochs. The training was done using NVIDIA RTX 4090 GPU.

\subsection{B.2~~Retrieval on NSD}

\quad \textbf{Model}. 
We use an MLP as the model backbone. It consists of a Voxel Projector and ResBlocks. The Voxel Projector comprises a sequence of operations: a Linear layer, a LayerNorm, a GLUE activation function, and a Dropout. It initially maps the voxel sequence into a latent space of 4096 dimensions. Following the above projection, the latent representations undergo further transformation within the latent space via 4 ResBlocks. Each ResBlock includes a Linear layer, a LayerNorm, a GLUE activation, a Dropout, and a residual connection from input to output. To achieve retrieval on NSD, we map 4096-dimension fMRI representation embedded by MLP to 257×768, the same as the image representation of CLIP ViT-L/14. Subsequently, we employ a Retrieval Projector that transfers the last dimension of fMRI representation. This Retrieval Projector consists of three Linear layers, sequentially expanding (768 to 2048), maintaining (2048 to 2048), and then reducing (2048 to 768) the dimension. The entire model is around 523M.

\textbf{Training}.
We set the batch size to 300 and use the AdamW optimizer. The learning rate linear is fixed at 3e-4 and the weight decay is fixed at 1e-4. For commonality learning, we set the weight of SRA loss to 0.05 and train 300 epochs. For fMRI pattern learning, we optimize the router for 10 epochs and then optimize the MoBEs for 10 meta steps. Each meta step contains training on the support set for 20 epochs and training on the query set for 5 epochs. The training was done using NVIDIA RTX 4090 GPU.

\subsection{B.3~~Few-Shot Learning}
For few-shot classification/retrieval, we adopt the same model and training hyperparameters as Appendix B.1/B.2.

\subsection{B.4~~Image Reconstruction on NSD}
\textbf{Model}. Please refer to Appendix A. The entire model is around 1.3B.
\textbf{Training}. We set the batch size to 32 and use the AdamW optimizer. We apply a linear warmup and a cosine learning rate scheduling from 1e-4 to 1e-5. The weight decay is fixed at 1e-4. For commonality learning, we set the weight of SRA loss to 0.05 and train 240 epochs. For fMRI pattern learning, we optimize the router for 10 epochs and then optimize the MoBEs for 20 meta steps. Each meta step contains training on the support set for 10 epochs and training on the query set for 5 epochs. The training was done using a single NVIDIA A100 80G GPU.

\subsection{B.5~~Evaluation Metrics}
We illustrate the evaluation metrics we adopt here.

\textbf{mAP}: The mean average precision metric refers to the average precision of all categories.

\textbf{AUC}: This metric refers to the area under the receiver operating characteristic curve.

\textbf{Hamming Distance}: This metric refers to the average of the number of mismatches between the predicted labels and the actual labels at each label position.

\textbf{Image Retrieval}: The Image Retrieval metric refers to retrieving the image representation with the highest cosine similarity based on fMRI representation on the test set (chance=1/300). If a paired image representation is retrieved, the retrieval is considered correct. 

\textbf{fMRI Retrieval}: The fMRI Retrieval metric refers to retrieving the fMRI representation with the highest cosine similarity based on image representation on the test set (chance=1/300). If a paired fMRI embedding is retrieved, the retrieval is considered correct. 

\textbf{PixCorr}: Pixel-wise correlation between ground truth image and reconstruction results.

\textbf{SSIM}: Structural similarity index metric~\cite{ssim} between ground truth and reconstructions .

\textbf{Eff}, \textbf{SwAV}: These two metrics refer to the average correlation distance with EfficientNet-B1~\cite{eff} and SwAV-ResNet50~\cite{swav}.

\textbf{Alex(2)}, \textbf{Alex(5)}, \textbf{Incep}, \textbf{CLIP}: all these metrics refer to two-way identification (chance = 50\%) using different models. Alex(2) denotes two-way comparisons are performed with the second layer of AlexNet, Alex(5) denotes comparisons with the fifth layer of AlexNet, Incep denotes comparisons with the last pooling layer of InceptionV3, and CLIP denotes comparisons with the final layer of ViT-L/14. Two-way identification refers to percent correct across comparisons gauging if the original image embedding is more similar to its paired voxel embedding or a randomly selected voxel embedding. We followed the same image preprocessing and the same two-way identification steps as previous works~\cite{minddiffuser, mindeye}.

\begin{table*}[t]
  \centering
  \scalebox{1.0}{
  \begin{tabular}{@{}ccccccccc@{}}
    \toprule
    \multirow{2}{*}{Subject} & \multirow{2}{*}{Architecture} && \multicolumn{3}{c}{Classification} & & \multicolumn{2}{c}{Retrieval} \\
    \cline{4-6} \cline{8-9}
    & && mAP$\uparrow$ & AUC$\uparrow$ & Hamming$\downarrow$ && Image $\uparrow$ & fMRI $\uparrow$  \\
    \midrule
    \multirow{3}{*}{Subj01}
        &  MLP           &&  0.476  &  0.949  &  0.0225 && 95.35\% & 83.90\%  \\
        &  Transformer   &&  0.428  &  0.941  &  0.0234 && 96.66\% & 86.06\% \\
        &  GFNet         &&  0.464  &  0.945  &  0.0234 && 95.26\% & 81.53\%  \\
    \midrule
    \multirow{3}{*}{Subj02}
        &  MLP           &&  0.437  &  0.942  &  0.0236 &&  95.35\%  &  83.90\%  \\
        &  Transformer   &&  0.401  &  0.932  &  0.0250 &&  95.21\%  &  81.63\%   \\
        &  GFNet         &&  0.422  &  0.938  &  0.0249 &&  95.01\%  &  80.35\% \\
    \midrule
    \multirow{3}{*}{Subj05}
        &  MLP          &&  0.507  &  0.955  &  0.0221  &&  83.94\%  &  68.84\% \\
        &  Transformer  &&  0.480  &  0.948  &  0.0235  &&  84.44\%  &  66.34\% \\
        &  GFNet        &&  0.478  &  0.950  &  0.0235  &&  89.97\%  &  67.19\% \\
    \midrule
    \multirow{3}{*}{Subj07}
        &  MLP          &&  0.448  &  0.933  & 0.0239 &&  84.35\%  &  71.12\% \\
        &  Transformer  &&  0.388  &  0.925  & 0.0247 &&  84.66\%  &  66.34\% \\
        &  GFNet        &&  0.406  &  0.932  & 0.0251 &&  85.99\%  &  68.70\%  \\
    \bottomrule
  \end{tabular}
  }
  \vspace{-1mm}
  \caption{Results of classification and retrieval on different model architectures. All results are averaged over three runs.}
  \label{table: robustness}
\end{table*}

\begin{table*}
 \centering
    \scalebox{0.94}{
    \begin{tabular}{@{}cccccccccc@{}}
    \toprule
Methods&\# Models&PixCorr$\uparrow$&SSIM$\uparrow$&Alex(2)$\uparrow$&Alex(5)$\uparrow$&Incep$\uparrow$&CLIP$\uparrow$&Eff$\downarrow$&SwAV$\downarrow$\\
    \midrule MindEye~\cite{mindeye}&4&\textbf{0.130}&0.308&\underline{84.0\%}&\underline{92.6\%}&86.9\%&86.1\%&0.778&0.477 \\
     Lite-Mind~\cite{lite-mind}&4&0.125&\underline{0.331}&78.7\%&89.4\%&\underline{87.9\%}&\underline{88.7\%}&\textbf{0.724}&\underline{0.446}\\
     \rowcolor{lightblue} Wills Aligner (\textbf{ours})&\textbf{1}&\underline{0.126}&\textbf{0.345}&\textbf{86.2}\%&\textbf{93.5\%}&\textbf{88.4}\%&\textbf{88.9\%}&\underline{0.730}&\textbf{0.441}\\
     \bottomrule
  \end{tabular}
  }
  \caption{Experimental results of the cross-modal retrieval on the LAION-5B. Our results are averaged over 4 subjects.}
  \label{table: LAION-5B retrieval}
\end{table*}

\section{C.~~Robustness for Model Architectures}
In the experiments described in the main text, we primarily used an MLP architecture as the backbone for the model $\mathcal{F}(\cdot)$. A key question is whether our Wills Aligner can be effectively generalized to other architectures. To investigate this, we extended our approach to different model architectures, specifically considering Transformers~\cite{transformer} and frequency domain learners like Lite-Mind~\cite{lite-mind} and GFNet~\cite{gfnet1, gfnet2}. We evaluated the robustness of the Wills Aligner using basic tasks such as classification and retrieval. If the model continues to extract visual information effectively after replacing the MLP with these alternative structures, it will demonstrate that our Wills Aligner is robust and can be generalized across different model architectures.

\subsection{C.1~~Architectures}

\begin{table}[t]
  \centering
  \scalebox{0.9}{
  \begin{tabular}{@{}ccccc@{}}
    \toprule
    \multirow{2}{*}{Architecture} & \multirow{2}{*}{Task} & \multicolumn{2}{c}{\# Parameters} \\
    \cline{3-4}
    & & Backbone & MoBE Adapters\\
    \midrule
     Transformer & Classification &  17.6M   &   1.6M (8.3\%) \\
     Transformer & Retrieval      &  35.9M   &   2.5M (6.5\%) \\
     GFNet & Classification       &  15.3M   &   1.8M (10.5\%) \\
     GFNet & Retrieval            &  33.7M   &   2.7M (7.4\%) \\
    \bottomrule
  \end{tabular}
  }
  \vspace{-1mm}
  \caption{Parameter counts of backbone and MoBE adapters in Transformer and GFNet.}
  \label{table: model parameter of Transformer and GFNet}
  \vspace{-2mm}
\end{table}

We adopt a Transformer and a frequency domain learner to replace MLP.  Table~\ref{table: model parameter of Transformer and GFNet} illustrates the model parameter counts of the Transformer and frequency domain learner. The following are details on model architectures.

\textbf{Transformer}. We employ the standard Vision Transformer. The voxel tokens and a CLS token undergo processing through 6 Transformer blocks. Each block comprises an 8-head Self-Attention layer alongside a FeedForward Network (FFN). Within the FFN, a layerNorm operation is initially applied, followed by an extension of the 768-dimensional tokens to 1024, subsequently reducing them back to 768 dimensions. In the classification task, the CLS token is directed to the Classifier Head, where it yields logits. For the retrieval task, we employ a Linear layer to expand the dimensionality of each non-CLS token to 8 times its original size, after which it's split into 8 tokens. This process results in a representation of size 257×768. 

\textbf{Frequency Domain Learner}.
Previous work~\cite{lite-mind} has shown that frequency-domain learners can effectively learn fMRI representation.
Therefore, we use GFNet~\cite{gfnet1, gfnet2} as our backbone network, which replaces the self-attention layers with frequency domain filters. We utilize 6 GFNet blocks to embed 33×768 fMRI representations. Subsequently, we employ the identical Classifier Head and Retrieval Projector for both classification and retrieval tasks.

\subsection{C.2~~Extended Data Preprocessing}
To cater to the input of Transformer and GFNet, we need to tokenize the 1-D voxel sequences. It can be formalized as:
\begin{equation}
    T_{s, i} = \mathrm{Conv}(\tilde{X}_{s, i}) + \mathrm{PE}~,
\end{equation}
where $\mathrm{PE}$ represent position encoding.
In the concrete implementation, we employ a 1-D convolution with an 1187 convolution kernel size and an 1187 stride for patchifying the sequences, adopting cosine position coding, thereby generating 32 voxel tokens. These voxel tokens $T_{s, i}$. These voxel tokens will be used as input to the model.

\subsection{C.3~~Results}
As illustrated in Figure~\ref{table: robustness}, different architectures demonstrate comparable visual decoding performance, indicating that our Wills Aligner is robust across various model architectures. Notably, the MLP architecture slightly outperforms the other two. We attribute this to the MLP's higher parameter count and the possibility of information loss in the tokenized voxels compared to the voxel sequences.

\section{D.~~Additional Results and Analysis}

\subsection{D.1~~Retrieval on LAION-5B}
Following MindEye, we performed large-scale fMRI-to-image retrieval in the LAION-5B dataset.

\textbf{Model}.
We first use the same MLP as retrieval on NSD. To achieve retrieval on LAION-5B, we map the 4096-dimension fMRI representation into 768, the same dimension as the CLS token of image representation. The entire model is around 147M parameters.

\textbf{Training}.
We set the batch size to 300 and use the AdamW optimizer. The learning rate linear descent from 1e-3 to 1e-5 and the weight decay is fixed at 1e-4. For commonality learning, we set the weight of SRA loss to 0.1 and train 300 epochs. For fMRI pattern learning, we optimize the router for 10 epochs and then optimize the MoBEs for 10 meta steps. Each meta step contains training on the support set for 20 epochs and training on the query set for 5 epochs. The training was done using NVIDIA RTX 4090 GPU.

\textbf{Results.} The results is shown in Table~\ref{table: LAION-5B retrieval}. It can be seen that our Wills Aligner achieves better results on most metrics, demonstrating the ability for large-scale retrieval.

\subsection{D.2~~Ablation on Rank of MoBE Adapters}

Since we use a low-rank decomposition for MoBE adapters, we intend to explore the effect of rank on the performance of our Wills Aligner. We use the MLP as the backbone architecture and perform classification and retrieval on the NSD.
We set the rank to [2, 4, 8, 16, 32]. Average results are reported for three different random seeds. 
As illustrated in Table~\ref{table: rank ablation}, the best performance is achieved when the rank is set to 16.
However, we observe that different ranks exhibit comparable performance, indicating that our method is insensitive to this hyperparameter.
This is also consistent with the previous conclusion of LoRA~\cite{lora}, which found that an increase in rank does not significantly improve performance.

\begin{table}[t]
  \centering
  \scalebox{0.9}{
  \begin{tabular}{@{}cccccccc@{}}
    \toprule
    \multirow{2}{*}{Rank} & & \multicolumn{3}{c}{Classifiaction} & & \multicolumn{2}{c}{Retrieval} \\
    \cline{3-5}
    \cline{7-8}
    & & mAP $\uparrow$ & AUC $\uparrow$ & Hamming $\downarrow$ & & Image $\uparrow$ & fMRI $\uparrow$ \\
    \midrule
    2  && 0.461 & 0.944 & \textbf{0.023} && 93.3\% & 80.5\% \\
    4  && 0.452 & 0.943 & \textbf{0.023} && 95.0\% & 82.4\% \\
    8  && 0.445 & 0.941 & 0.024 && 93.9\% & 82.6\% \\
    16 && \textbf{0.467} & \textbf{0.945} & \textbf{0.023} && \textbf{95.4\%} & 83.9\% \\
    32 && 0.448 & 0.942 & 0.024 && 95.2\% & \textbf{84.7\%} \\
    \bottomrule
  \end{tabular}
  }
  \vspace{-2mm}
  \caption{Experiments for ranks of MoBE adapters. Results are the average of three runs.} 
  \label{table: rank ablation}
  \vspace{-2mm}
\end{table}

\subsection{D.3~~Runtime comparisons on NSD}
we conducted runtime comparisons on NSD in terms of inference time (ms) per sample and FLOPs (B). The results is shown in Table~\ref{runtime}. The results show our method is faster on the classification and retrieval tasks and slightly slower on reconstruction, indicating that it is computationally efficient on large datasets.

\begin{table}[t]
  \centering
  \scalebox{0.9}{
  \begin{tabular}{@{}cccccc@{}}
    \toprule
    \multirow{2}{*}{Task} & \multicolumn{2}{c}{Best Baselines} && \multicolumn{2}{c}{Wills Aligner} \\
    \cline{2-3}
    \cline{5-6}
    & Inference & FLOPs && Inference & FLOPs \\
    \midrule
     Classification & 21 & 2.0 && 12 & 0.6 \\
     Retrieval      & 4.3 & 6.9 && 3.6 & 2.5 \\
     Reconstruction & 201.1 & 560.9 && 203.5 & 561.3 \\
     \bottomrule
  \end{tabular}
  }
  \vspace{-1mm}
  \caption{Result of runtime comparisons on NSD.}
  \label{runtime}
  \vspace{-2mm}
\end{table}

\subsection{D.4~~Additional Ablation on Anatomical Alignment}

In order to investigate the role of anatomical alignment, we simulate fMRI misalignment through performing random Voxel Dropout to obtain the same fMRI length of 12k, and then conduct ablation studies on the NSD classification task. The results are shown in Table~\ref{alignment ablation}. It can be seen that misalignment will result in significant performance degradation.

\begin{table}[t]
  \centering
  \begin{tabular}{@{}ccc@{}}
    \toprule
    Method & mAP$\uparrow$ & AUC$\uparrow$ \\
    \midrule
     Anatomical Alignment & 0.424 & 0.937 \\
     Misalignment         & 0.247 & 0.854 \\
     \bottomrule
  \end{tabular}
  \vspace{-1mm}
  \caption{Ablation study of anatomical alignment.}
  \label{alignment ablation}
  \vspace{-2mm}
\end{table}

\subsection{D.5~~Retrieval Results Visualisation}
\begin{figure*}[h]
  \centering
  \includegraphics[width=0.55\linewidth]{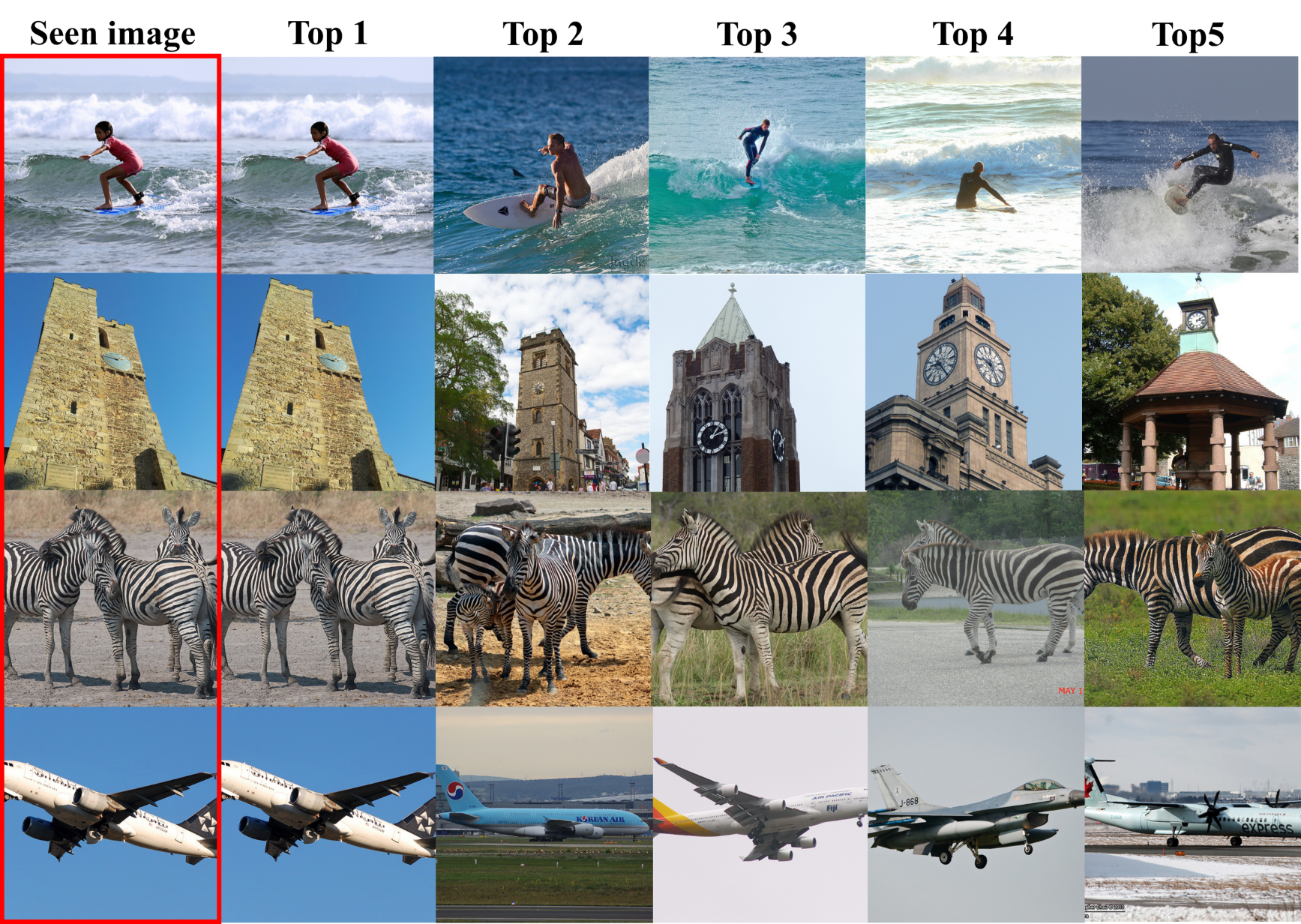}
   \caption{Retrieval results of Subj01 on the test set.}
   \label{retri1}
\end{figure*}

\begin{figure*}[h]
  \centering
  \includegraphics[width=0.67\linewidth]{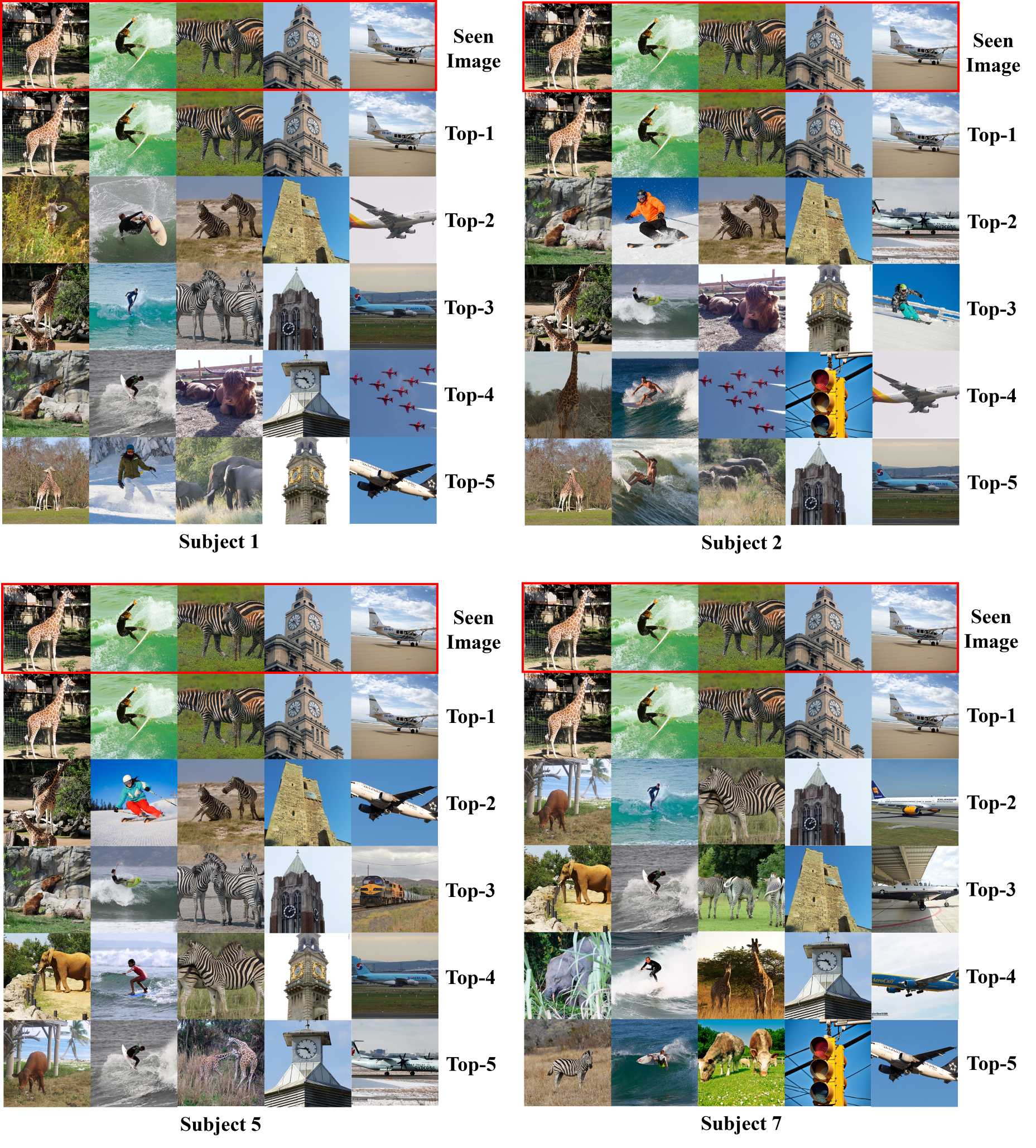}
   \caption{Retrieval results of all four subjects on the test set.}
   \label{retri2}
\end{figure*}

\begin{figure*}[h]
  \centering
  \includegraphics[width=0.55\linewidth]{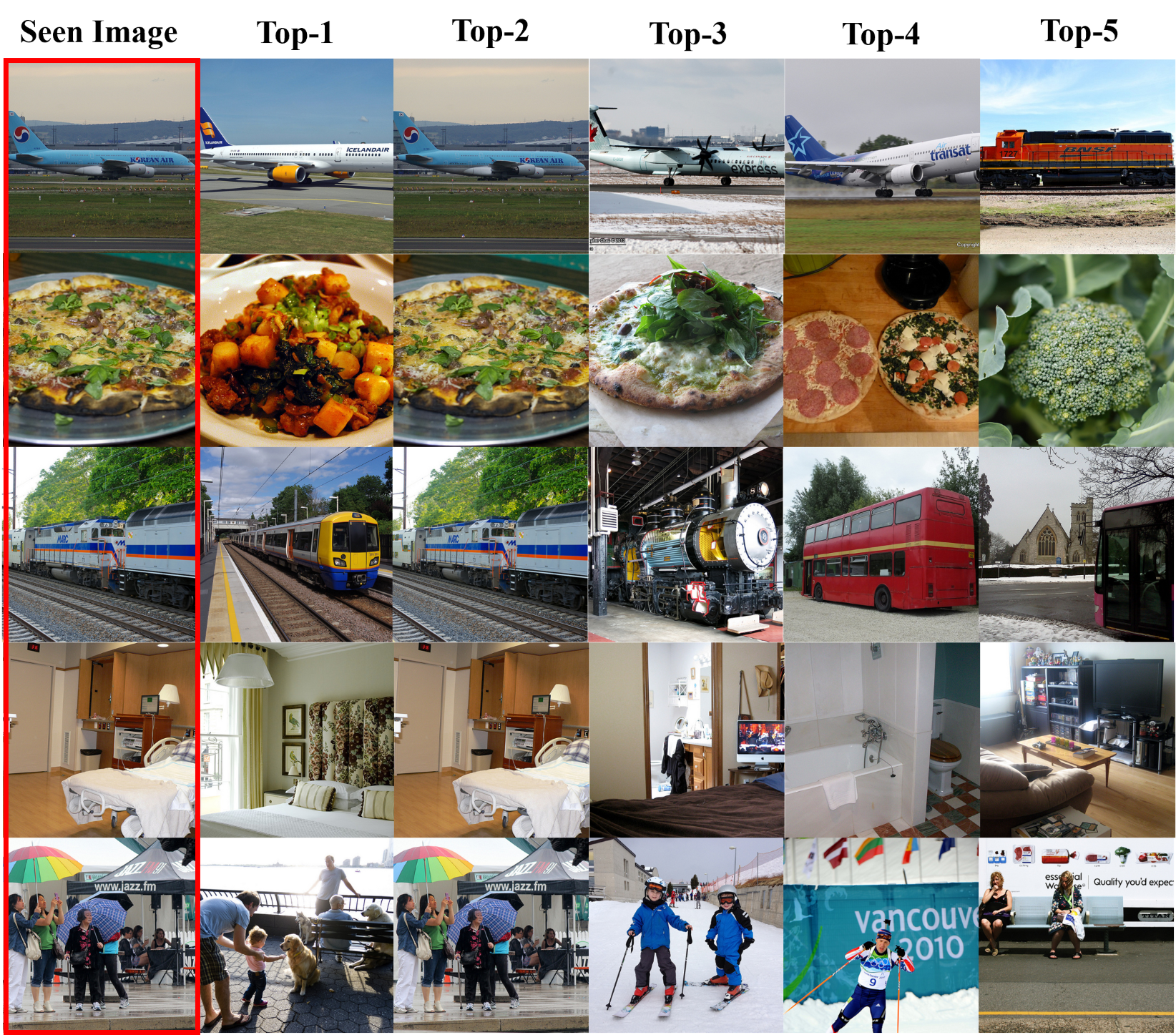}
   \caption{Failed retrieval results of Subj01 on the test set.}
   \label{retri3}
\end{figure*}

Figure~\ref{retri1} and Figure~\ref{retri2} show the result of top-5 image retrieval on the entire testing set. 
The results demonstrate the effectiveness of our Wills Aligner in accurately retrieving images according to fMRI. Furthermore, we observe that the top-5 retrieval results consistently exhibit high semantic relevance.
However, our model also makes mistakes, as shown in Figure~\ref{retri3}.
We observe that incorrect image retrievals often occur with semantically rich images featuring complex scenes. The brain's visual cortex struggles to accurately extract complete semantics from these highly intricate images, leading to the loss of detailed information in the fMRI and resulting in errors. Additionally, the proportion of semantically irrelevant images increases in the top 2-5 retrieval results. We believe this is also influenced by the composition of the retrieval pool. In the NSD test set of 982 samples, images depicting zebras, trains, giraffes, surfing, skiing, architecture, airplanes, and similar subjects are more common, whereas images of food, crowds, furniture, and similar subjects are less frequent. This imbalance reduces the semantic relevance of the top 2-5 retrieval results.

\subsection{D.6~~Additional Result of Image Reconstruction}
\begin{figure*}[h]
  \centering
  \includegraphics[width=0.75\linewidth]{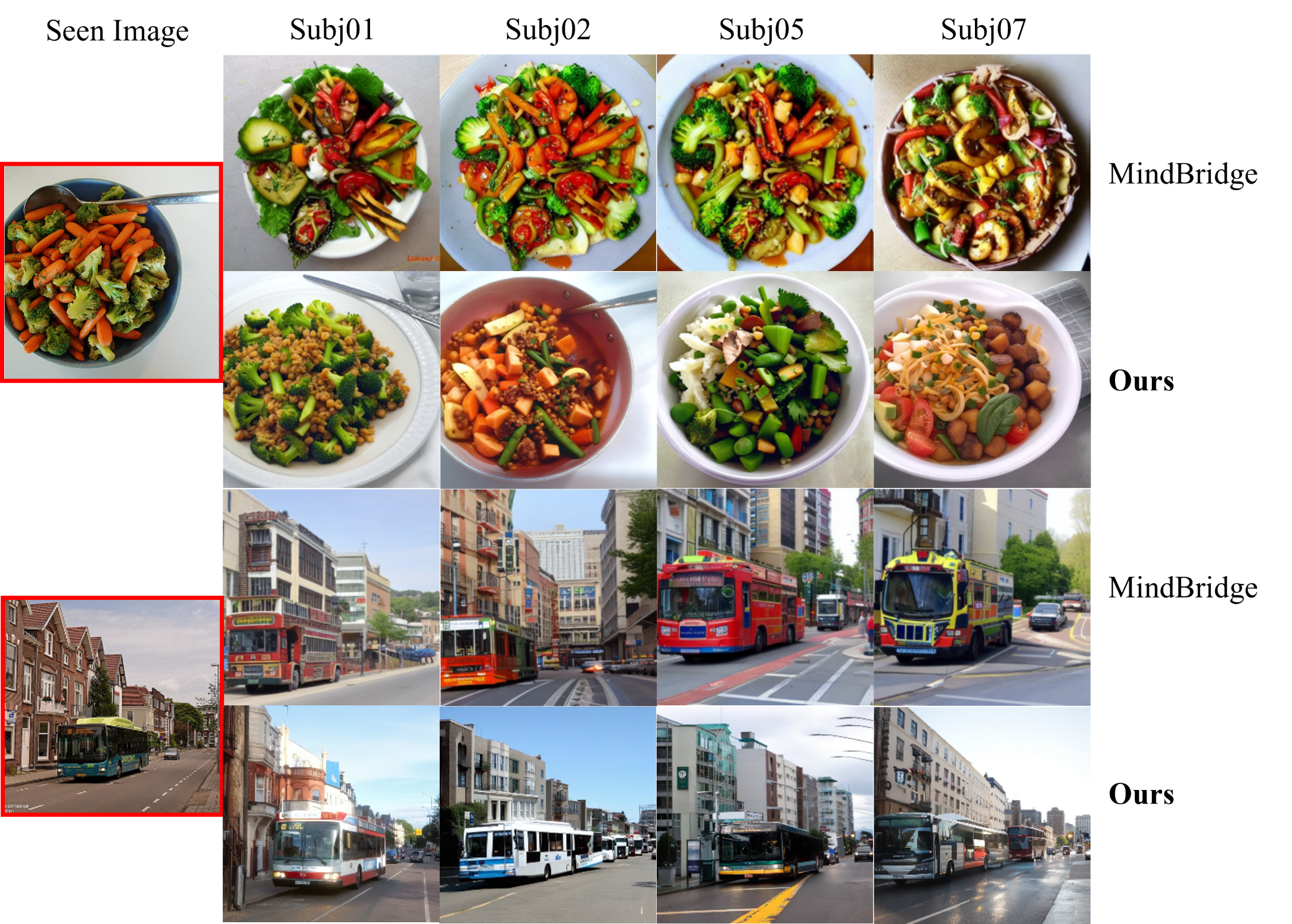}
   \caption{Reconstruction results compared with MindBridge.}
   \label{recon2}
\end{figure*}

\begin{figure*}[h]
  \centering
  \includegraphics[width=0.63\linewidth]{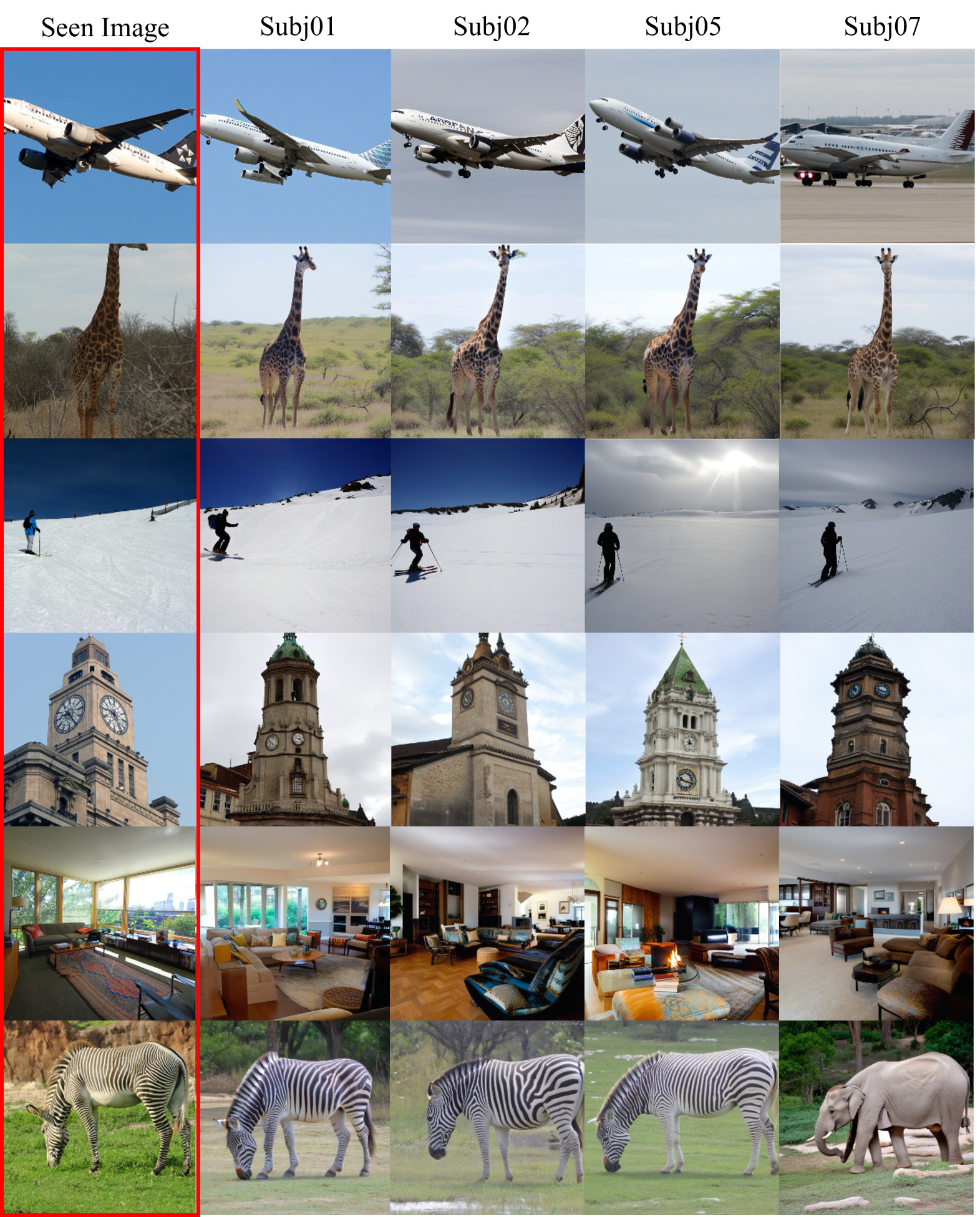}
   \caption{More reconstruction results of our Wills Aligner.}
   \label{recon3}
\end{figure*}

\begin{figure*}[h]
  \centering
  \includegraphics[width=0.63\linewidth]{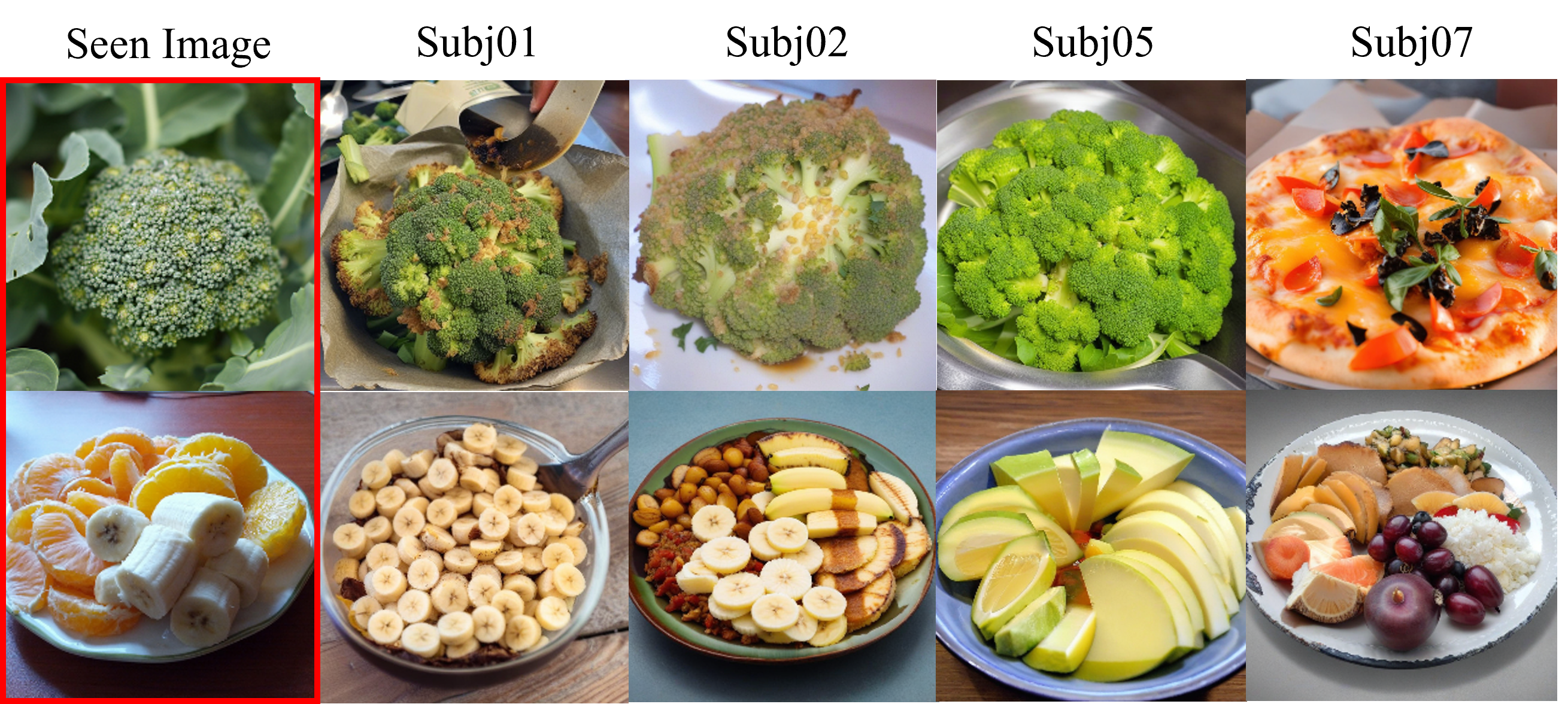}
   \caption{Failed reconstruction results of our Wills Aligner.}
   \label{recon4}
\end{figure*}

Figure~\ref{recon2} shows the comparison of our reconstruction results with MindBridge. First, we correctly reconstruct the contents of the Seen Image and the reconstruction is superior to MindBridge, containing more details. Second, our reconstruction will vary more between subjects due to the fact that our MoBE adapters learn the fMRI patterns of individual subjects.
Figure~\ref{recon3} shows more results of our reconstruction, which is of high quality and faithful to the seen images.
Figure~\ref{recon4} shows the results of some failed reconstructions. For example, subj04 is not able to reconstruct the food “broccoli” in the top row. The fidelity of the reconstruction in the bottom row is not sufficient, and items are often reconstructed beyond what is in the seen image. We believe that the lack of fidelity is a consequence of the prior knowledge of the generative model, which is currently a major problem in visual reconstruction.

\section{E.~~Datasets}
\subsection{E.1~~Basic Information}
The Natural Scenes Dataset (NSD)~\cite{nsd} is a massive 7T neuroscience dataset encompassing fMRI data from 8 subjects. Throughout the NSD experiment, participants were presented with images sourced from MSCOCO~\cite{mscoco}, while their neural responses were recorded utilizing a high-resolution 7-Tesla fMRI scanner. Our study aligns with previous research~\cite{mindeye,xia2024dream,mindbridge,lite-mind} by concentrating on Subj01, Subj02, Subj05, and Subj07, as these 4 subjects completed all 40 session scans. Within the dataset, each subject's train set comprises 8859 distinct visual stimuli and 24980 fMRI, wherein each image was repetitively presented for viewing 1-3 times. The test set contains 982 visual stimuli and 2770 fMRI. It is crucial to emphasize that within the training set for each subject, the visual stimuli employed do not overlap with those utilized for other subjects. However, the same visual stimuli are used across subjects within the test set. Upon selecting the region-of-interest (ROI) named \textit{nsdgeneral}, we acquired voxel sequences for the four subjects, whose lengths are 15724, 14278, 13039, and 12682, respectively.

\subsection{E.2~~Detials on Anatomical Alignment}
We map the fMRI data from different subjects to a standard brain template, called \textit{fsaverage7}. This brain template is a triangular surface mesh created via the spherical registration of 40 individuals using an energy minimization algorithm to align surface-based features~\cite{fsaverage}. FreeSurfer provides tools facilitating the alignment of any brain to this brain template. After anatomical alignment, selecting ROI, and spatial flattening, the 1-D voxel sequences are standardized to a uniform voxel length of 37984.

\subsection{E.3~~Ethical Statement}
All the datasets we use are open-source datasets and do not have ethical issues.

\clearpage

\end{document}